# A Genetic Algorithm for the Multi-Pickup and Delivery Problem with time windows

**I. Harbaoui Dridi**[(1),(2)]   **R. Kammarti**[(1),(2)]   **M. Ksouri**[(2)]   **P. Borne**[(1)]
imenharbaoui@gmail.com   kammarti.ryan@planet.tn   Mekki.Ksouri@insat.rnu.tn   p.borne@ec-lille.fr

[(1)] LAGIS : Ecole Centrale de Lille, Villeneuve d'Ascq, FRANCE
[(2)] LACS : Ecole Nationale des Ingénieurs de Tunis, Tunis - Belvédère. TUNISIE

**Abstract**: In This paper we present a genetic algorithm for the multi-pickup and delivery problem with time windows (m-PDPTW). The m-PDPTW is an optimization vehicles routing problem which must meet requests for transport between suppliers and customers satisfying precedence, capacity and time constraints. This paper purposes a brief literature review of the PDPTW, present our approach based on genetic algorithms to minimizing the total travel distance and thereafter the total travel cost, by showing that an encoding represents the parameters of each individual.

## 1. Introduction

With the time and economic constraints implications of the transport goods problem, its resolution becomes very difficult, requiring the use of tools from different disciplines (manufacturing, information technology, combinatorial optimization, etc.)... Indeed, the process from transport systems and scheduling are becoming more complex by their large size, by the nature of their relationship dynamics, and by the multiplicity of which they are subjected.

Many studies have been directed mainly towards solving the vehicle routing problem (VRP). It's an optimization vehicle routing problem to meet travel demands. Other researchers became interested on an important variant of VRP which is the PDPTW (Pickup and Delivery Problem with Time Windows) with capacity constraints on vehicle.

The PDPTW is divided into two: 1-PDPTW (single-vehicle) and m-PDPTW (multi-vehicle).

Our object is to design a tool for m-PDPTW resolution based on genetic algorithms and Pareto dominance method to give a set of satisfying solutions to this problem minimizing the total travel distance and thereafter the total travel cost, by showing that an encoding represents the parameters of each individual.

## 2. Literature review

### 2.1 Vehicle routing problem

The Vehicle Routing Problem (VRP) represents a multi-goal combinatorial optimization problem which has been the subject of much work and many variations in the literature. It belongs to the NP-hard class. . [Christofides, N and al., 1979] [Lenstra, J and al., 1981]

Heuristic methods have been used by Matsinen, T and al to solve the general problem in this area as the Travelling Salesman Problems [Matsinen, T and al., 1992]. The Meta heuristics were as well applied to solving the vehicle routing problem. Among these methods, we can include ant colony algorithms, which were used by Montamenni, R and al for the resolution of DVRP and used also by Sorin C, N and al to solving the multiple constraints problem: vehicle route allocation. [Montamenni, R and al., 2002] [Sorin C, N and al., 2008]

The VRP principle is: given a depot D and a set of customers orders C = (c1, ... , Cn), to build a package routing, for a finite number of vehicles, beginning and ending at a depot. In these routing, a customer must be served only once by a single vehicle and vehicle capacity transport for a routing should not be exceeded. [Nabaa, M and al., 2007]

Savelsbergh shown that the VRP is a NP-hard problem [Savelsbergh, M.P.W and al., 1995]. Since the m-PDPTW is a generalization of the VRP it's a NP-hard problem.

**2.2 The PDPTW: Pickup and Delivery Problem with Time Windows**

The PDPTW is a variant of VRPTW where in addition to the existence of time constraints, this problem implies a set of customers and a set of suppliers geographically located. Every routing must also satisfy the precedence constraints to ensure that a customer should not be visited before his supplier. [Psaraftis, H.N., 1983]

A dynamic approach for resolve the 1-PDP without and with time windows was developed by Psaraftis, H.N considering objective function as a minimization weighting of the total travel time and the non-customer satisfaction. [Psaraftis, H.N., 1980]

Jih, W and al have developed an approach based on the hybrid genetic algorithms to solve the 1-PDPTW, aiming to minimize combination of the total cost and total waiting time. [Jih, W and al., 1999]

Another genetic algorithm was developed by Velasco, N and al to solve the 1-PDP bi-objective in which the total travel time must be minimized while satisfy in prioritise the most urgent requests. In this literature, the method proposed to resolve this problem is based on a No dominated Sorting Algorithm (NSGA-II). [Velasco, N and al., 2006]

Kammarti, R and al deal the 1-PDPTW, minimizing the compromise between the total travel distance, total waiting time and total tardiness time, using an evolutionary algorithm with Special genetic operators, tabu search to provide a set of viable solutions. [Kammarti, R and al., 2004] [Kammarti, R and al 2005a]
This work have been extended, in proposing a new approach based on the use of lower bounds and Pareto dominance method, to minimize the compromise between the total travel distance and total tardiness time. [Kammarti, R and al 2006] [Kammarti, R and al 2007]

About the m-PDPTW, Sol, M and al have proposed a branch and price algorithm to solve the m-PDPTW, minimizing the vehicles number required to satisfy all travel demands and the total travel distance. [Sol, M and al., 1994]

Quan, L and al have presented a construction heuristic based on the integration principle with the objective function, minimizing the total cost, including the vehicles fixed costs and travel expenses that are proportional to the travel distance. [Quan, L and al., 2003]

A new metaheuristic based on a tabu algorithm, was developed by Li, H and al to solve the m-PDPTW. [Li, H and al., 2001]

Li, H and al have developed a "Squeaky wheel" method to solve the m-PDPTW with a local search. [Li, H and al., 2002]

A genetic algorithm was developed by Harbaoui Dridi, I and al dealing the m-PDPTW to minimize the total travel distance and the total transport cost. [Harbaoui Dridi, I and al., 2008]

## 3. Mathematical formulation

Our problem is characterized by the following parameters:

- $N$ : Set of customers, supplier and depot vertices,
- $N'$: Set of customers and supplier vertices,
- $N^+$ : Set of supplier vertices,
- $N^-$ : Set of customers vertices,
- $K$ : Vehicle number,
- $d_{ij}$ : Euclidian distance between the vertex $i$ and the vertex $j$. If $d_{ij} = \infty$ then the road between $i$ and $j$ doesn't exist,
- $t_{ijk}$ : Time used by the vehicle $k$ to travel from the vertex $I$ to the vertex $j$,
- $[e_i, l_i]$ : Time window of the vertex $i$,
- $s_i$ : Stopping time at the vertex $i$,
- $q_i$ : Goods quantity of the vertex $i$ request. If $q_i > 0$, the vertex $i$ is a supplier; if $q_i < 0$, the vertex $i$ is a customer and if $q_i = 0$ then the vertex was served.
- $Q_k$ : Capacity of vehicle $k$,
- $i = 0..N$ : Predecessor vertex index,
- $j = 0..N$ : Successor vertex index,
- $k$: 1..K: Vehicle index,
- $X_{ijk} = \begin{cases} 1 & \textit{If the vehicle travel from the vertex i to the vertex j} \\ 0 & \textit{Else} \end{cases}$
- $A_i$ : Arrival time of the vehicle to the vertex $i$,
- $D_i$ : Departure time of the vehicle from the vertex $i$,
- $y_{ik}$ : The goods quantity in the vehicle $k$ visiting the vertex $I$,
- $C_k$ : Travel cost associated with vehicle k,
- A vertex is served only once,
- There is one depot,
- The capacity constraint must be respected,
- The depot is the start and the finish vertex for the vehicle,
- The vehicle stops at every vertex for a period of time to allow the request processing,
- If the vehicle arrives at a vertex $i$ before its time windows beginning date $e_i$, it waits.

The function to minimize is given as follows:

$$Minimiser\ f = \left( \sum_{k \in K} \sum_{i \in N} \sum_{j \in N} C_k d_{ij} X_{ijk} \right) \qquad (1)$$

Subject to:

$$\sum_{i=1}^{N} \sum_{k=1}^{K} x_{ijk} = 1,\ j = 2,...N \qquad (2)$$

$$\sum_{j=1}^{N} \sum_{k=1}^{K} x_{ijk} = 1,\ i = 2,...N \qquad (3)$$

$$\sum_{i \in N} X_{i0k} = 1, \forall k \in K \tag{4}$$

$$\sum_{j \in N} X_{0jk} = 1, \forall k \in K \tag{5}$$

$$\sum_{i \in N} X_{iuk} - \sum_{j \in N} X_{ujk} = 0, \forall k \in K, \forall u \in N \tag{6}$$

$$X_{ijk}=1 \Rightarrow y_{jk} = y_{ik} + q_i, \forall i,j \in N; \forall k \in K \tag{7}$$

$$y_{0k}=0, \forall k \in K \tag{8}$$

$$Q_k \geq y_{ik} \geq 0, \forall i \in N; \forall k \in K \tag{9}$$

$$D_w \leq D_v, \forall i \in N; \forall w \in N_i^+; \forall v \in N_i^- \tag{10}$$

$$D_0 = 0 \tag{11}$$

$$X_{ijk} = 1 \Rightarrow e_i \leq A_i \leq l_i, \forall i,j \in N; \forall k \in K \tag{12}$$

$$X_{ijk} = 1 \Rightarrow e_i \leq A_i + s_i \leq l_i, \forall i,j \in N; \forall k \in K \tag{13}$$

$$X_{ijk} = 1 \Rightarrow D_i + t_{ijk} \leq (l_j - s_j), \forall i,j \in N; \forall k \in K \tag{14}$$

The constraint (2) and (3) ensure that each vertex is visited only once by a single vehicle. The constraint (4) and (5) ensure that the vehicle route beginning and finishing is the depot. The constraint (6) ensures the routing continuity by a vehicle.

(7), (8) and (9) are the capacity constraints. The precedence constraints are guaranteed by (10) and (11). The constraints (12), (13) and (14) ensure compliance time windows.

### 4. Genetic algorithm for minimizing the total travel cost

In this part, we present our approach based on genetic algorithms to minimize the total travel cost.

### 4.1 Solution coding

A chromosome is a succession (permutation) vertex, which indicates the order in which a vehicle is to visit all the vertices. Fig.1 represents the solutions under form of chromosomes.

| Vertex (i) | 0 | 5 | 8 | 2 | 6 | 4 | 3 | 10 | 7 | 9 | 1 | 0 |

**Fig.1: Solution coding**

The vertex "0" represents the depot.

### 4.2 Decoding the vehicles passage

The decoding allows, from every chromosome of the population to obtain an initial solution sign the passage of each vehicle on the corresponding vertex. (Fig.2)

| $V_1$ | $C_1$ | 0 | 2 | 4 | 1 | 5 | 0 |
|---|---|---|---|---|---|---|---|
| $V_2$ | $C_2$ | 0 | 3 | 6 | 7 | 10 | 0 |
| $V_3$ | $C_3$ | 0 | 8 | 9 | 0 | | |

**Fig.2: Order of vehicles passage**

### 4.3 Crossover operator

Following the generation of the initial population, we proceed to crossover phase which ensures the recombination of parental genes for train new descendants. To do this, we choose the one point crossover. (Fig.3)

| 1 | 2 | 3 | 4 | 5 |   | 6 | 7 | 8 | 9 | 10 |
|---|---|---|---|---|---|---|---|---|---|---|

$P=2$

| 1 | 2 | 8 | 9 | 10 |   | 6 | 7 | 3 | 4 | 5 |
|---|---|---|---|---|---|---|---|---|---|---|

**Fig.3 : Crossover operator**

### 4.4 Mutation operator

Mutation operator (Fig.4) is to choose two positions at random, within a chromosome and exchange their respective values.

| 1 | 2 | 3 | 4 | 5 |
|---|---|---|---|---|

| 4 | 2 | 3 | 1 | 5 |
|---|---|---|---|---|

**Fig.4 : Mutation operator**

Note here that we must respect the constraint precedence, to ensure that a customer is not visited before his supplier, and capacity to ensure the non-overload of each vehicle.

### 4.5 Procedure of the proposed approach for minimizing the total travel cost

#### 4.5.1 Generation of initial population

The choice of the initial population is important because it can make a genetic algorithm more or less fast to converge towards the global optimum.

In our case, we will generate two types of populations. A first population noted $P_{node}$, which represents all nodes to visit with all vehicles, according to the permutation list coding (Fig.1). The second population noted $P_{vehicle}$ indicates nodes number visited by each vehicle. Knowing that $k$ varies between 1 and $\frac{N'}{2}$ vehicles. Fig.5 shows an individual example of $P_{vehicle}$ with $N' = 10$.

| $V_1$ | $V_2$ | $V_3$ | $V_4$ | $V_5$ |
|---|---|---|---|---|
| 6 | 4 | 0 | 0 | 0 |

Fig.5: Individual example of $P_{vehicle}$

### 4.5.2 Correction procedure

The principle of correction precedence and capacity is to ensure that a customer is not visited before his supplier while respecting the vehicles capacity.

Whereas couples customer / supplier following: (1.5), (2.8), (9.7), (10.3) and (4.6), noting that $Q_{k\,max} = 60$, we present, respectively, in Fig.6 and Fig.7 the principle of correction precedence and capacity.

| 0 | 3 | 2 | 6 | 8 | 1 | 4 | 5 | 9 | 10 | 7 | 0 |

*Before correction*

| 0 | 3 | 8 | 2 | 6 | 5 | 1 | 4 | 7 | 9 | 10 | 0 |

*After correction*

**Fig.6 : Correction precedence**

| 0 | 5 | 8 | 7 | 3 | 1 | 2 | 4 | 9 | 6 | 10 | 0 |

$q=80$

*Before correction*

| 0 | 5 | 8 | 7 | 9 | 3 | 1 | 2 | 4 | 6 | 10 | 0 |

*After correction*

**Fig.7 : Correction capacity**

### 4.5.3 Generation of population $P_{node/vehicle}$

Considering the population $P_{node}$ given by Fig.1, correction procedures and $P_{vehicle}$ population, given by Fig.5, we illustrated in Fig.8 an individual of the population $P_{node/vehicle}$.

| $V_1$ | $C_1$ | 0 | 5 | 8 | 2 | 6 | 4 | 3 | 0 |
| $V_2$ | $C_2$ | 0 | 10 | 7 | 9 | 1 | 0 | | |

**Fig.8: Individual of the population $P_{node/vehicle}$**

### 4.5.3 Calculation procedure

Each vehicle will serve vertices in the assignment order. Once returned to the depot, we calculate the transport cost corresponding to the travel distance, and we repeat this until all the vertices are served.

We reproduce this work for each individual to the population $P_{vehicle}$, taking into account the different possible combinations between $P_{vehicle}$ and $P_{node}$, to obtain thereafter the correspondent individual of the population $P_{node/vehicle}$ that minimizes our objective function. The algorithm of this procedure is shown by the Fig.9.

Fig.10 indicates how to minimizing the total cost of transport.

```
Begin
Step 1 : Create the initial population, (size n).
Step 2 : Fill the intermediate population $P_{node}$ (size 2n) with individuals'
crossover, mutation or copy.
Step 3 : Correction procedure of Precedence and capacity.
Step 4 : Create the 2$^{nd}$ intermediate population $P_{node/vehicle}$ (size 2n * 2n)
representing the routing of each vehicle.
While the generation number is not reached do
Step 5 : Determine fitness values for each individual of the
population $P_{node/vehicle}$ .(travel cost)
Copy the best solution
If the current solution is better then
        The best solution = current solution
        End If
Increment generations number
End
End
```

**Fig.9: Approach algorithm**

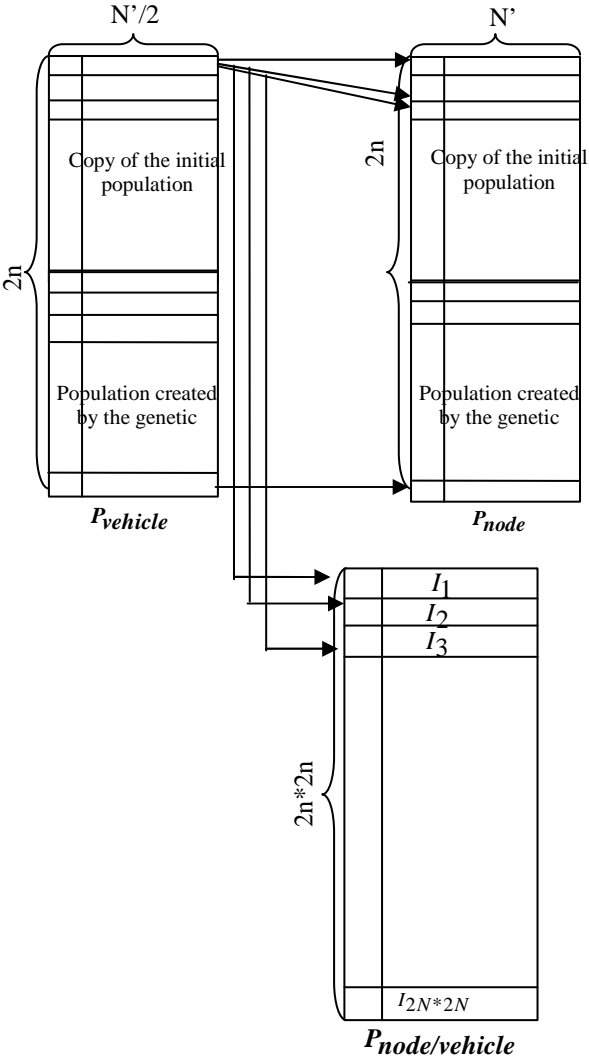

**Fig. 10: Calculation Procedure**

With:

n: size of initial population,

$f_i$ : objective function to minimize (given by equation 1) for iteration i, i = 1 ... 2n * 2n.

We determine thereafter:

$$f_{research} = \min_{i} f_i \qquad (15)$$

### 4.5.4 Simulation

We present in Table 1, the simulation results given by our approach.

| $N'$ | $n$ | $k$ | $Min\left(\sum_{k \in K} \sum_{i \in N} \sum_{j \in N} d_{ijk} X_{ijk}\right)$ | $\min_{i} f_i$ |
|---|---|---|---|---|
| 20 | 100 | 2 | 878,84 | 59425,95 |
| 20 | 500 | 2 | 843,39 | 52296,84 |
| 50 | 100 | 2 | 2379,22 | 145534,61 |
| 50 | 500 | 2 | 2234,06 | 140743,17 |
| 100 | 100 | 4 | 4581,24 | 265197,12 |
| 100 | 500 | 4 | 4439,64 | 257383,70 |

**Table 1 : Simulation results**

We observe that our approach reduces the vehicles number (2 vehicles to serve up to 50 vertices and 4 vehicles to serve 100 vertices).
Knowing that a vehicle will cause an augmentation travel cost on the addition of a driver, the travel fixed costs, to depreciation of the vehicle used. Hence the interest of our approach that aims to minimize the total travel cost.

### 5. Conclusion

In this paper we have proposed, following a study of the different approaches proposed for the resolution of 1-PDPTW and m-PDPTW, a genetic algorithm for minimizing the total travel cost. To do this, we presented the mathematical formulation of our problem, and we have detailed the calculation procedure for determine the optimal solution, that minimizes our objective function

**Références bibliographiques**